\documentclass[sigconf]{acmart}
\AtBeginDocument{%
  }

\setcopyright{acmlicensed}
\copyrightyear{2018}
\acmYear{2018}
\acmDOI{XXXXXXX.XXXXXXX}
\acmConference[Conference acronym 'XX]{Make sure to enter the correct
  conference title from your rights confirmation email}{June 03--05,
  2018}{Woodstock, NY}
\acmISBN{978-1-4503-XXXX-X/2018/06}

\begin{document}

\settopmatter{authorsperrow=5}

\title{A Unified AI Approach for Continuous Monitoring of Human Health and Diseases from Intensive Care Unit to Home with Physiological Foundation Models (UNIPHY+)}


%\affiliation{%
%  \institution{Emory University}
%  \city{Atlanta}
%  \state{GA}
%  \country{USA}
%}
%\email{}

%\affiliation{%
%  \institution{Children’s Healthcare of Atlanta}
%  \city{Atlanta}
%  \state{GA}
%  \country{USA}
%}
%\email{}

%\affiliation{%
%  \institution{Nihon Kohden Digital Health Solutions, Inc}
%  \city{Irvine}
%  \state{CA}
%  \country{USA}
%}
%\email{}

\author{Minxiao Wang}
\authornote{Both authors contributed equally to this research.}
\affiliation{%
  \institution{Emory University}
  \city{Atlanta}
  \state{GA}
  \country{USA}
}
\email{mwang80@emory.edu}

\author{Saurabh Kataria}
\authornotemark[1]
\affiliation{%
  \institution{Emory University}
  \city{Atlanta}
  \state{GA}
  \country{USA}
}
\email{skatar6@emory.edu}

\author{Juntong Ni}
\affiliation{%
  \institution{Emory University}
  \city{Atlanta}
  \state{GA}
  \country{USA}
}
%\email{juntong.ni@emory.edu}
%\email{jni44@emory.edu}

\author{Tim Buchman}
\affiliation{%
  \institution{Emory Healthcare}
  \city{Atlanta}
  \state{GA}
  \country{USA}
}
%\email{tbuchma@emory.edu}

\author{Jocelyn~Grunwell}
\authornote{Full affiliation: Children’s Healthcare of Atlanta} 
\affiliation{%
  %\institution{Children’s Healthcare of Atlanta}
  \institution{CHOA}
  \city{Atlanta}
  \state{GA}
  \country{USA}
}
%\email{jocelyn.grunwell}
%\email{@choa.org}
%\email{jgrunwe@emory.edu}

\author{Mark Mai}
\authornotemark[2]
\affiliation{%
  %\institution{Children’s Healthcare of Atlanta}
  \institution{CHOA}
  \city{Atlanta}
  \state{GA}
  \country{USA}
}
%\email{mark.mai@choa.org}

\author{Wei Jin}
\affiliation{%
  \institution{Emory University}
  \city{Atlanta}
  \state{GA}
  \country{USA}
}
%\email{wei.jin@emory.edu}
%\email{wjin30@emory.edu}

\author{Matthew Clark}
\authornote{Full affiliation: Nihon Kohden Digital Health Solutions, Inc} 
\affiliation{%
  %\institution{Nihon Kohden Digital Health Solutions, Inc}
  \institution{NKDHS Inc}
  \city{Irvine}
  \state{CA}
  \country{USA}
}
%\email{matthew_clark}
%\email{@nihonkohden.com}

\author{Stephanie~Brown}
\authornotemark[2]
\affiliation{%
  %\institution{Children’s Healthcare of Atlanta}
  \institution{CHOA}
  \city{Atlanta}
  \state{GA}
  \country{USA}
}
%\email{stephanie.brown}
%\email{@choa.org}

\author{Michael Fundora}
\authornotemark[2]
\affiliation{%
  %\institution{Children’s Healthcare of Atlanta}
  \institution{CHOA}
  \city{Atlanta}
  \state{GA}
  \country{USA}
}
%\email{fundoram@kidsheart.com}

\author{Puneet Sharma}
\affiliation{%
  \institution{Emory University}
  \city{Atlanta}
  \state{GA}
  \country{USA}
}
%\email{puneet.sharma2@emory.edu}
%\email{pshar41@emory.edu}

\author{Tony Pan}
\affiliation{%
  \institution{Emory University}
  \city{Atlanta}
  \state{GA}
  \country{USA}
}
%\email{tony.pan@emory.edu}
%\email{tcpan@emory.edu}

\author{Sam Khan}
\affiliation{%
  \institution{Emory Healthcare}
  \city{Atlanta}
  \state{GA}
  \country{USA}
}
%\email{sharaf.khan@emory.edu}
%\email{skhan90@emory.edu}

\author{Timothy Ruchti}
\authornotemark[3]
\affiliation{%
  %\institution{Nihon Kohden Digital Health Solutions, Inc}
  \institution{NKDHS Inc}
  \city{Irvine}
  \state{CA}
  \country{USA}
}
%\email{timothy_ruchti}
%\email{@nihonkohden.com}

\author{Naveen Muthu}
\affiliation{%
  \institution{Emory University}
  \city{Atlanta}
  \state{GA}
  \country{USA}
}
%\email{nmuthu2@emory.edu}

\author{Kevin Maher}
\authornotemark[2]
\affiliation{%
  %\institution{Children’s Healthcare of Atlanta}
  \institution{CHOA}
  \city{Atlanta}
  \state{GA}
  \country{USA}
}
%\email{maherk@kidsheart.com}

\author{Siva Bhavani}
\affiliation{%
  \institution{Emory Healthcare}
  \city{Atlanta}
  \state{GA}
  \country{USA}
}
\email{sbhava2@emory.edu}

\author{Xiao Hu}
\affiliation{%
  \institution{Emory University}
  \city{Atlanta}
  \state{GA}
  \country{USA}
}
\email{xiao.hu@emory.edu}
\renewcommand{\shortauthors}{Wang and Kataria et al.}  % Short header

\begin{abstract}
We present UNIPHY+, a unified physiological foundation model (physioFM) framework designed to enable continuous human health and diseases monitoring across care settings using ubiquitously obtainable physiological data. We propose novel strategies for incorporating contextual information during pretraining, fine-tuning, and lightweight model personalization via multi-modal learning, feature fusion-tuning, and knowledge distillation. We advocate testing UNIPHY+ with a broad set of use cases from intensive care to ambulatory monitoring in order to demonstrate that UNIPHY+ can empower generalizable, scalable, and personalized physiological AI to support both clinical decision-making and long-term health monitoring.
\end{abstract}




\received{20 February 2007}
\received[revised]{12 March 2009}
\received[accepted]{5 June 2009}

\maketitle

\section{Introduction}

Sensor data of human physiology ($\sim$ physiological data) enable proactive detection of critical health changes from patient monitoring in intensive care units (ICUs) to home monitoring of chronic conditions using wearable sensors. With the increased use of wearables in other medical settings and at home, the scope of physiological monitoring is immense and the potential for improving human health at scale is unprecedented. However, physiological data are prone to noise and artifacts of a multi-scale and nonstationary nature and hence are difficult to analyze. For example, the high number of false alarms \cite{1} from standard physiological monitors limits their impact on early warning of deterioration. While deep learning is increasingly used to process physiological data, mainstream approaches design a specific model for a specific use case. This paradigm is difficult to scale to handle diverse use cases especially for those with limited amount of labelled training data-a situation that is often encountered in both inpatient and ambulatory settings.

Recently, the concept of foundation model (FM) has emerged \cite{2}. The success of pretraining large transformers with internet-scale text as a starting point to develop specific solutions for a variety of downstream tasks has demonstrated the power of FM. Following this recipe for developing FMs for physiological data ($\sim$ physioFM) is a promising and logical next step \cite{3,4,5,6,7,8,9,10}. We recognize \cite{11} that human physiology is interconnected by the nervous, circulatory, endocrine, and immunological systems. Hence, a physiological signal directly reflecting one organ function contains information about other organs \cite{12}, leading to our premise that a well-trained physioFM may capture this intrinsic physiological interdependencies and unlock patient monitoring capabilities using ubiquitously obtainable signals like ECG and photoplethysmography (PPG).

A generative pretrained transformer (GPT)-based physioFM was trained by us using 2.6 million hours of PPG data from 25,000 adult ICU patients (though from a single institution) with a model size ranging from 19M to 1B parameters. This PPG FM was fine-tuned to obtain state-of-the-art results in physiological measurement (PMEA) tasks including heart rate (HR), respiration rate (RR), blood pressure, and atrial fibrillation (AF) detection \cite{13}. In an ongoing work, this PPG FM has shown promising results to represent raw PPG to predict in-hospital cardiopulmonary arrest (IHCA) achieving an area under the receiver operator characteristic curve of \textbf{0.8153} and an area under the precision-recall curve of \textbf{0.1177} at an IHCA prevalence of 5\%.

Physiological data are typically treated as time series by AI researchers who explored large language models (LLMs) for time series \cite{14,15,16}. These approaches have intrinsic limitations due to LLM’s poor ability to encode numerical data \cite{17}. Hence, others adapted the successful recipe of pretraining LLMs to train time series FMs from scratch including Chronos \cite{18}, MOMENT \cite{19}, PatchTST \cite{20}. They showed some promising results in analyzing physiological signals though with limited testing. This cross-domain generalizability of time series FMs motivates growing efforts towards building specialist FMs pretrained with physiological waveforms. ECG and PPG signals are the most relevant to developing a ubiquitous monitoring solution. There are a variety of model architectures and training methods proposed for ECG FMs \cite{21,22,23,24,25}. However, none of these models incorporated contextual information such as subjects’ organ function status in pretraining. PPG signals are even more ubiquitous and can be passively and continuously acquired using over-the-counter commercial devices with little burden on patients. In addition to our GPT-based PPG FM, we explored convolutional neural network as the backbone and a novel contrastive loss to pretrain a PPG FM. The loss function was based on a physiological insight that two PPG recordings obtained within a short time window (e.g., five minutes) will likely represent similar physiological states.5 Researchers from Apple also reported a PPG FM but only tested the model on a limited set of static tasks such as estimating ages \cite{3}. PaPaGEI \cite{8} incorporated PPG pulse morphology to design a contrastive loss and used a fully connected neural network as a backbone. Though, it was pretrained with only 57,000 hours of data. More recently, another PPG FM \cite{27} was proposed where a distance function to compare two PPG signals was learned based on systematically comparing motifs between two PPG signals, but only using data from 120 patients. Again, none of these existing PPG FMs use contextual information about the data during pretraining.

Existing physioFMs have not reached the scale of LLMs in terms of both the model size and the amount of pretraining data. For example, our PPG FM was trained with roughly 2.5 million hours of data. If we treat a single heartbeat PPG as a token, our model was merely trained with billions of tokens – much smaller than trillions of tokens used in pretraining LLMs. In addition, contextual information such as organ status and treatments are important to help interpret physiological data. Yet, such information has not been included in pretraining physioFMs. Furthermore, such information may be critical when fine-tuning a physioFM for specific tasks particularly before a large-scale physioFM becomes available. Because physioFM is positioned to enhancing human health monitoring with wearable data from individual subjects. In theory, physiological data can be obtained nonintrusively 24 $\times$ 7 resulting in billions of physiological tokens even for an individual person. Techniques to distill and personalize a physioFM with an ever-increasing amount of data per subject are thus needed. We believe that incorporating contextual information to pre-train and fine-tune physioFM, as well as leveraging individual subject’s own data to personalize models, are areas for further innovations to realize the full potential of physioFM and we will outline some of the ideas in the next section.

\section{Further Innovations to Build and Apply PhysioFM}

The following three innovative ideas cover all three critical stages of applying a physioFM to solve specific tasks: pretraining, fine-tuning, and model personalization.

We note that most of the physiological data to pre-train physioFM will likely be sourced from inpatients with rich electronic health records (EHRs). To incorporate contextual information, we propose to treat a sequence of physiological data as a source “language” and a sequence of organ functions and treatment as captured in its matched EHR data as a target “language” to pretrain FMs using a translator architecture. Pretraining a physioFM in this way essentially incorporates pathophysiological contextual information that is missing in pretraining approaches that only use physiological data \cite{3,4,5,6,7,8,9,10}. In addition to the representation power of a physioFM, models thus pretrained can be used to create digital twins to simulate and investigate treatment effects leveraging the generative nature of the model.

\begin{figure}[htbp]
  \includegraphics[width=\linewidth]{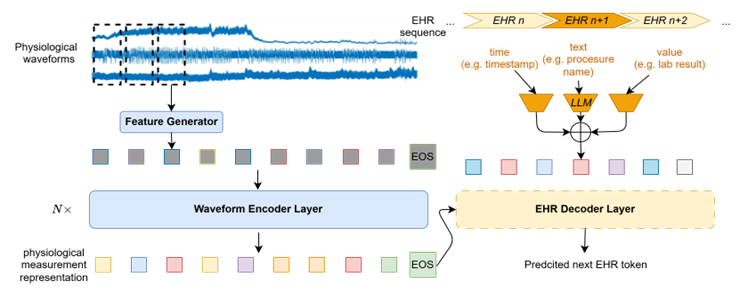}
  \caption{Translator architecture to pretrain a physioFM with EHR tokens.}
  \label{fig2}
\end{figure}

One way to implement this idea is illustrated in Fig.~\ref{fig2} using an encoder-decoder architecture. After pretraining, the encoder will be a physioFM that captures contextual association, in a temporally causal way, between intrinsic dynamics in waveform and organ functions and responses to treatments as captured in EHR data. Such a model would be able to represent physiological data with more general and broader knowledge and ultimately make fine-tuning for specific tasks easier. In our design, the encoder and decoder backbones are standard Transformer-based architectures as adopted in Google T5 \cite{28}. However, we need to introduce new tokenization approaches for waveforms and EHR data. First, we will process waveforms to derive feature vectors at a coarser time scale, e.g., five minutes. These features will be derived using signal processing approaches as implemented in open-source toolboxes including pyPPG \cite{29} and neuroKit2 \cite{30}. To tokenize EHR data, we will adopt a modified triplet approach \cite{31}. %We will focus on lab test results, procedures, and medications. As illustrated in Fig.\ref{fig2}, a sequence of physiological features will be appended with a special token [EOS] to indicate the end of a physiological signal strip. The learned embedding of an EOS token will be used at every step of the decoder to form cross-attention with EHR tokens and predict the next EHR token in an autoregressive fashion. 

\begin{figure}[htbp]
  \includegraphics[width=0.8\linewidth]{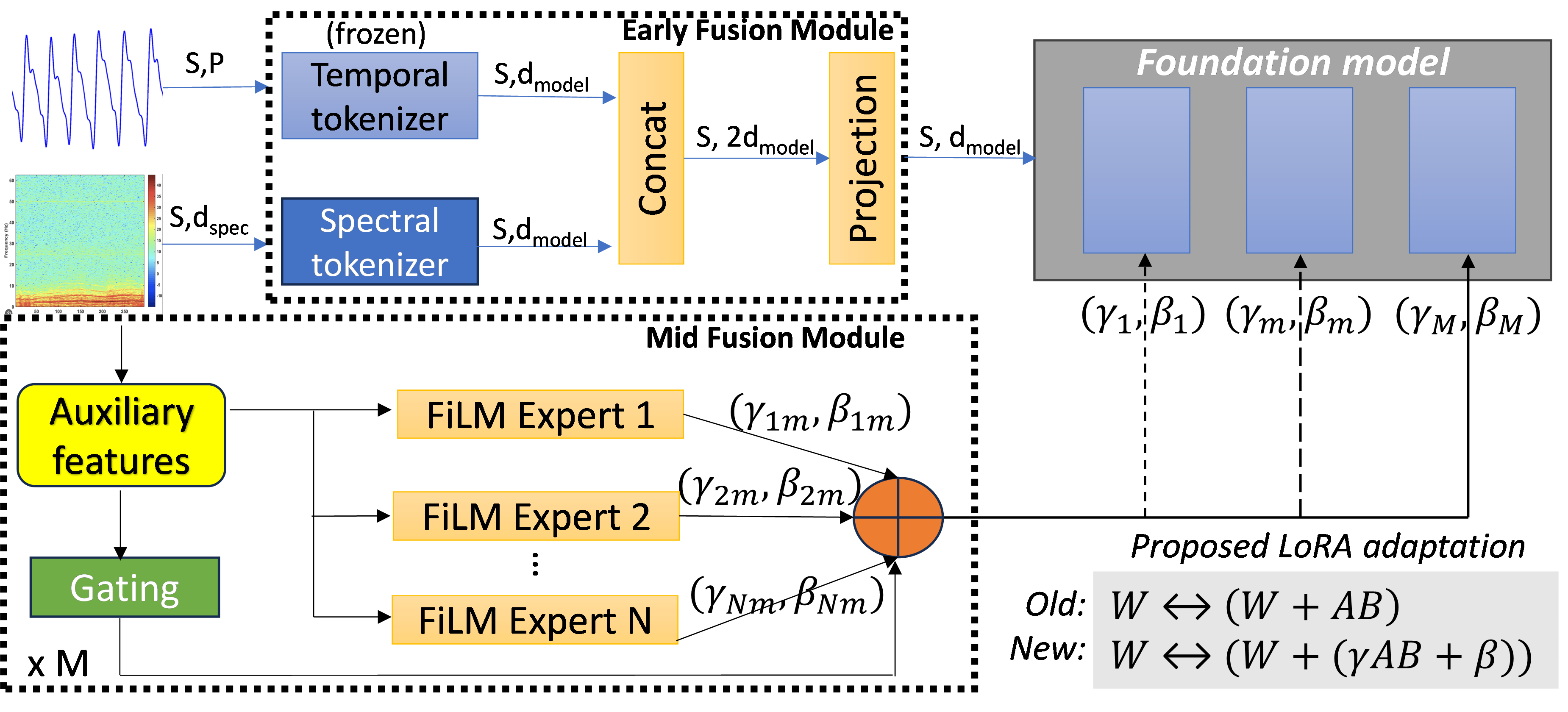}
  \caption{Feature-based fine-tuning of foundation models.}
  \label{fig3}
\end{figure}

We propose a novel feature fusion-tuning algorithm that allows the use of additional features to fine-tune a pre-trained FM in order to specialize with richer task-specific data. For example, prediction of asthma exacerbation using PPG/ECG could be enhanced with incorporation of respiratory sound signals. As illustrated in Fig. \ref{fig3}, fuse-tuning has two configurations: 1) Early fusion (or reprogramming \cite{32}) happens in the first layer of the model. This fusion of tokens can be further enhanced with more sophisticated techniques like gating-based fusion as illustrated in the figure \cite{33}. 2) For fusing extra features in the middle of FM, we propose to extend the established LoRA algorithm \cite{34}, which learns two low-rank matrices which are added to the frozen weight matrix of a pretrained FM. We repurpose the original formulation with a conditional version, which can accept extra features. To maximize the benefit of our approach, we add a mixture-of-expert sub-model that accepts extra features as input and independently devises the affine parameters.

\begin{figure}[htbp]
  \includegraphics[width=0.8\linewidth]{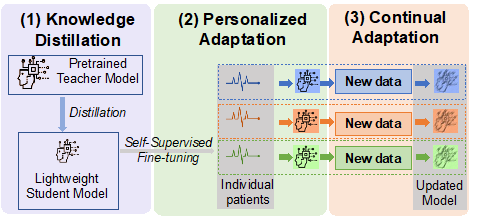}
  \caption{Three components of physioDistill.}
  \label{fig4}
\end{figure}

We propose to tackle model personalization from angles of model distilling and continual learning to sustain optimal performance for individual subjects while reducing size of a physioFM. A general physioFM faces two limitations: (1) it is computationally intensive for real-time, on-device deployment, and (2) it may be further optimized for the unique physiological characteristics and health trajectories of individual subjects. We propose PhysioDistill (Fig.\ref{fig4}), which designs a novel knowledge distillation framework \cite{35,36} that derives compact, high-performing models tailored to individuals from a physioFM. PhysioDistill includes three interconnected components: (1) Model Compression via Knowledge Distillation; (2) Personalized Adaptation with Self-Supervised Objectives (3) Continual Adaptation with New Patient Data or disease conditions.

\section{Data, Experiments, and Expected Results}
Physiological data from wearable devices are rarely collected and archived in the same systematic way that electronic health records and device data from inpatient settings are routinely captured. Therefore, we believe that in foreseeable future, inpatient physiological data from routine bedside patient monitors with matched EHRs will be the primary source of data for pretraining physioFMs. However, a key step moving forward is to collate such data from more than one institution. An ongoing effort under NIH’s Bridge2AI program to build a flagship critical care dataset for AI was conceived before the era of ChatGPT but its relevance for developing foundation models is now even more critical.

%Physiological data from wearable devices are rarely collected as systematically as inpatient data from bedside monitors and EHRs. Thus, in the near term, inpatient physiological data with matched EHRs will likely serve as the primary source for pretraining physioFMs. A critical next step is to aggregate such data across institutions. Notably, the NIH Bridge2AI initiative, launched before the rise of foundation models like ChatGPT, is now highly relevant for this purpose.

Because a key success criterion for physioFM is the demonstration of its versatility in performing diverse downstream tasks, a study of physioFMs will need to be broad and cover use cases across different patient care settings from ICUs to homes. In ICU settings, tracking and prediction of dynamic changes of patients prior to subacute events such as sepsis and acute cardiorespiratory failures are clinically important use cases. In ambulatory settings, using physioFM to extract, from routine signals such as PPG and ECG, more biological and physiological markers of human health will be impactful to help develop a stronger and more informative feedback loop to guide healthy behavior changes. Plausible biomarkers that physioFM has the potential estimate would include glucose, certain electrolytes, and even lactates. This area is far less studied and would need some initial exploration with proper study designs and data collection, but the proposed algorithmic elements can well support such exploration studies.

%To demonstrate the versatility of physioFMs, studies must cover diverse care settings, from ICUs to home environments. In ICUs, key use cases include tracking and predicting dynamic patient changes preceding events such as sepsis or acute cardiorespiratory failure. In ambulatory settings, physioFMs could extract meaningful health markers, like glucose, electrolytes, and lactate, from routine signals like PPG and ECG, offering feedback to support healthy behavioral changes. Although understudied, this area is ripe for exploration using the proposed algorithmic framework.

Ultimately, we expect that the pursuit of a generalizable and contextualized physioFM will advance patient monitoring technology in both inpatient and ambulatory settings. Multi-parameter physiological monitoring is ubiquitous in acute care settings but data collected are under-utilized to support clinical decision-making. Therefore, it is imperative to leverage the use of continuous physiological signals to improve the accuracy, timeliness, and generalizability of clinical AI solutions. Physiological patient monitoring can now be performed with new technologies such as multi-parameter patches and wearables \cite{41,42}. Large industry studies \cite{43, 44, 45} have demonstrated the scale at which human health and diseases such as AF can be monitored and detected by using commercial devices with AI-powered algorithms. It should be noted that these prior studies were not using the latest AI approaches. With the proposed physioFM to further unlock physiological measurement capabilities in these signals to track organ functions through continuously estimating electrolytes, lactate, glucose, and other organ function indicators, the potential to use such information to close the loop to precisely guide individual behavior changes to promote healthy diet, exercise, sleep quality and to achieve these objectives at scale is unprecedented \cite{46}.

%We anticipate that developing a generalizable, contextualized physioFM will enhance patient monitoring across care environments. While multi-parameter monitoring is routine in acute care, its data remain underutilized for decision-making. Leveraging continuous physiological signals can improve the accuracy, timeliness, and generalizability of clinical AI tools. Emerging technologies, such as wearable patches, enable scalable physiological monitoring \cite{41,42}, and large-scale industry studies \cite{43,44,45} have shown the feasibility of AI-powered health tracking. With the proposed physioFM to further unlock physiological measurement capabilities in these signals to track organ functions through continuously estimating electrolytes, lactate, glucose, and other organ function indicators, the potential to use such information to close the loop to precisely guide individual behavior changes to promote healthy diet, exercise, sleep quality and to achieve these objectives at scale is unprecedented \cite{46}.

%\begin{acks}
%.
%\end{acks}

\bibliographystyle{ACM-Reference-Format}
\bibliography{UNIPHY_references}

\end{document}